\documentclass[twoside]{article}

\usepackage[accepted]{aistats2026}
\usepackage[round]{natbib}

\usepackage{amsfonts}
\usepackage{algorithm}
\usepackage{algpseudocode}
\usepackage{booktabs}
\usepackage{multirow}
\usepackage{graphicx}
\usepackage{float}
\usepackage{graphicx}
\usepackage{subcaption}
\usepackage{xcolor}
\usepackage{hyperref}
\usepackage[noabbrev,capitalise,nameinlink]{cleveref}
\hypersetup{
    colorlinks=true,
    linkcolor=[RGB]{0, 0, 139},    % Dark Blue
    citecolor=[RGB]{0, 100, 0}     % Dark Green
}

\begin{document}

\runningtitle{Adaptive Replay Buffer (ARB)}


\twocolumn[

% \aistatstitle{Instructions for Paper Submissions to AISTATS 2026}
\aistatstitle{Adaptive Replay Buffer for\\ Offline-to-Online Reinforcement Learning}

% \aistatsauthor{ Author 1 \And Author 2 \And  Author 3 }
\aistatsauthor{ Chihyeon Song \And Jaewoo Lee \And  Jinkyoo Park }

\aistatsaddress{ KAIST \And KAIST, MongooseAI \And KAIST, Omelet } ]

\begin{abstract}
Offline-to-Online Reinforcement Learning (O2O RL) faces a critical dilemma in balancing the use of a fixed offline dataset with newly collected online experiences. Standard methods, often relying on a fixed data-mixing ratio, struggle to manage the trade-off between early learning stability and asymptotic performance. To overcome this, we introduce the Adaptive Replay Buffer (ARB), a novel approach that dynamically prioritizes data sampling based on a lightweight metric we call 'on-policyness'. Unlike prior methods that rely on complex learning procedures or fixed ratios, ARB is designed to be learning-free and simple to implement, seamlessly integrating into existing O2O RL algorithms. It assesses how closely collected trajectories align with the current policy's behavior and assigns a proportional sampling weight to each transition within that trajectory. This strategy effectively leverages offline data for initial stability while progressively focusing learning on the most relevant, high-rewarding online experiences. Our extensive experiments on D4RL benchmarks demonstrate that ARB consistently mitigates early performance degradation and significantly improves the final performance of various O2O RL algorithms, highlighting the importance of an adaptive, behavior-aware replay buffer design. Our code is publicly available at \url{https://github.com/song970407/ARB}.
\end{abstract}

\section{Introduction}

Offline Reinforcement Learning (Offline RL) trains policies exclusively from a fixed, pre-collected dataset \citep{levine2020offline, prudencio2023survey}. This approach circumvents the need for costly and potentially unsafe real-world interactions, making it highly valuable in scenarios where data collection is challenging \citep{gu2023human, zeng2024survey}. In real-world scenarios, boosting a policy's performance with newly acquired data from online interaction is often necessary. However, Offline RL shows an extremely slow rate of performance gain due to its conservative learning approach, which is designed to prevent catastrophic extrapolation errors \citep{kumar2020conservative, kumar2019stabilizing, kostrikov2021offline}. To address this and to find a way to effectively leverage valuable data from online interactions, the Offline-to-Online Reinforcement Learning (O2O RL) paradigm has emerged.

O2O RL combines the stability of pre-training on an offline dataset with the adaptability of fine-tuning through online interaction \citep{zheng2023adaptive}. The process typically involves training an initial policy using offline data and then progressively updating it through interaction with the environment. This fine-tuning allows the policy to adapt to novel dynamics and achieve superior asymptotic performance. However, a central challenge in this paradigm is the replay buffer dilemma: how to effectively compose the replay buffer with a mixture of both offline and newly collected online data. Naive methods, such as relying solely on offline data, can cap performance \citep{nakamoto2023cal}, while using only new online data can lead to catastrophic forgetting and performance collapse \citep{nair2020awac, kostrikov2021offline}. This highlights the need for a dynamic and principled method for mixing data. While prior work has attempted to address this with methods such as a fixed data-mixing ratio \citep{ball2023efficient} or by learning a metric for online-ness of experiences \citep{lee2022offline}, these solutions suffer from significant drawbacks. A fixed ratio is often sub-optimal across different environments, and learning an additional metric introduces computational overhead and algorithmic complexity.

% To illustrate this dilemma, consider the two naive extremes of data-mixing. Relying solely on the offline dataset provides a stable, conservative learning signal but prevents the policy from adapting to novel states and actions discovered online, thus capping performance \citep{nakamoto2023cal}. Conversely, fine-tuning with only new online data can lead to catastrophic forgetting of previously acquired knowledge from pretraining and a subsequent performance collapse. The limited and often out-of-distribution nature of early online data can destabilize the value function, leading to a severe performance drop \citep{nair2020awac, kostrikov2021offline}. This highlights the need for a dynamic and principled method for mixing data. While prior work has attempted to address this with methods such as a fixed data-mixing ratio \citep{ball2023efficient} or by learning a metric for online-ness of experiences  \citep{lee2022offline}, these solutions suffer from significant drawbacks. A fixed ratio is often sub-optimal across different environments, and learning an additional metric introduces computational overhead and algorithmic complexity.

In this paper, we propose a simple yet effective solution to the dilemma of the replay buffer. We introduce the Adaptive Replay Buffer (ARB), a novel strategy for a replay buffer that dynamically prioritizes data sampling based on a lightweight metric. Our metric measures how likely a data point is to have been generated by the current policy, effectively capturing its online-ness. A key strength of ARB is its simplicity and versatility; it can be seamlessly integrated into most existing O2O RL algorithms without requiring any additional models or complex training procedures. We demonstrate through extensive experiments on D4RL benchmarks that integrating ARB consistently mitigates initial performance degradation and significantly boosts the final performance of state-of-the-art O2O RL agents.

\section{Related Works}
\subsection{Offline-to-Online Reinforcement Learning}

Offline RL enables policy training from fixed datasets by addressing the challenge of estimation drift. To mitigate the risk of overestimating values for out-of-distribution (OOD) actions, standard approaches employ estimation optimization techniques, such as conservative value learning \citep{kumar2020conservative} or in-sample learning \citep{kostrikov2021offline}. While these methods ensure stability within the static dataset, this inherent conservatism often becomes a bottleneck during online fine-tuning. The strict constraints required for safe offline learning can inhibit the exploration needed to discover higher-reward regions in the dynamic online environment, resulting in slow performance gains \citep{lee2022offline, yu2023actor, nakamoto2023cal, ball2023efficient}.
%rebuttal로 추가됨
To overcome this limitation and effectively leverage the pre-trained prior, the Offline-to-Online Reinforcement Learning (O2O RL) paradigm has emerged. O2O RL aims to combine the stability of offline pre-training with the adaptability of online interaction, managing the delicate trade-off between retaining offline knowledge and adapting to distributional shifts. However, this approach faces significant challenges, particularly a performance drop that can occur when an agent transitions from the static offline dataset to a dynamic online environment. To overcome this, many methods have been proposed to manage the distributional shift and the tendency of a policy to select OOD actions.
%rebuttal 이전 버전
% O2O RL is a promising paradigm for improving the efficiency and performance of RL agents by combining offline pre-training with online fine-tuning. However, this approach faces significant challenges, particularly a performance drop that can occur when an agent transitions from the static offline dataset to a dynamic online environment. To overcome this, many methods have been proposed to manage the distributional shift and the tendency of a policy to select out-of-distribution (OOD) actions.

\subsubsection{Policy-based Methods}
One key strategy involves using policy constraints during the online fine-tuning phase to keep the learned policy close to the data distribution. For instance, AWAC \citep{nair2020awac} leverages an implicit policy constraint, while Adaptive Behavior Cloning Regularization \citep{zhao2022adaptive} dynamically adjusts the constraint strength based on the agent's performance. Similarly, Adaptive Policy Learning \citep{zhao2022adaptive} modifies the policy learning objective to better adapt to online data. Another recent approach, FamO2O \citep{wang2023train}, moves beyond a single, fixed constraint by training a family of policies with varying improvement-constraint balances and then adaptively selecting the most suitable policy for each state based on online feedback.
\subsubsection{Value-Based and Generative Methods}
Another effective approach is to optimize how offline and online data are used for training by modifying the value function or generating new samples. Cal-QL \citep{nakamoto2023cal} addresses the value overestimation problem directly by calibrating the Q-values during fine-tuning, which helps prevent policy collapse. The Actor-Critic Alignment method \citep{yu2023actor} addresses a misalignment that can occur between the pre-trained actor and the critic, which can cause performance degradation during the transition from offline to online learning. More recently, novel approaches have emerged, such as Energy-guided Diffusion Sampling \citep{liu2024energy}, which uses a generative model to create new samples that conform to the desired online data distribution.
\subsubsection{Other Data-Centric Methods}
Other approaches provide a foundational framework for optimizing the use of offline and online data. Methods like Jump-Start RL \citep{uchendu2023jump} provide a foundational framework for using a prior policy to guide online exploration, significantly accelerating learning in the initial stages. SPOT \citep{wu2022supported} introduces a density-based regularization to explicitly model the behavior policy's support, ensuring a smooth transition to online learning. Additionally, Policy Expansion \citep{zhang2023policy} is a distinct method that tackles fine-tuning by adding a new, learnable policy to an existing offline policy set, allowing for exploration without degrading the original policy. These diverse approaches highlight the ongoing effort to create stable and efficient O2O RL algorithms that can effectively leverage pre-collected data to accelerate learning in a target environment.
\subsection{Replay Buffer in Reinforcement Learning}
The replay buffer is an essential component in off-policy RL algorithms for enhancing data efficiency and reducing sample correlations \citep{mnih2013playing, fedus2020revisiting}. While traditional replay buffers sample uniformly, Prioritized Experience Replay \citep{schaul2015prioritized} improved learning efficiency by prioritizing samples based on "importance", such as TD error. Beyond simple prioritization, other advanced strategies have been proposed to make experiences more useful, such as Hindsight Experience Replay \citep{andrychowicz2017hindsight}, which learns from failed experiences by relabeling them with new goals. Furthermore, research has explored optimizing the replay buffer sampling distribution itself for improved performance \citep{zha2019experience}.

In the context of O2O RL, Balanced Experience Replay Buffer \citep{lee2022offline} enhances sample efficiency by prioritizing a mix of online and offline samples that leverages the "onlineness" of the data. Methods like Emphasizing Recent Experience \citep{wang2019boosting} have also been developed to strategically prioritize recently collected data, which is especially important for balancing new and old knowledge in the buffer. Warm-start RL \citep{zhou2024efficient} demonstrates that efficient fine-tuning is possible without the need to retain offline data by recalibrating the offline Q-function to online data through a small number of warmup rollouts.

While BERB also prioritizes onlineness, it does so by learning an additional metric, which adds computational overhead and algorithmic complexity. In contrast, our Adaptive Replay Buffer (ARB) achieves a similar goal without requiring an additional learning process or model. ARB is a learning-free and lightweight solution that directly uses the policy's log-likelihood as a proxy for on-policyness, making it a more versatile and efficient plug-in solution. This distinction is critical as it simplifies implementation and reduces computational costs, allowing for seamless integration into a wide range of O2O RL algorithms.
\section{Preliminaries}
\subsection{Reinforcement Learning and Markov Decision Process}

The goal of an RL agent is to learn the optimal policy $\pi^*: \mathcal{S}\rightarrow \mathcal{A}$ for a given Markov Decision Process $\mathcal{M}=(\mathcal{S}, \mathcal{A}, P,r,\rho_0, \gamma)$, consisting of a state space $\mathcal{S},$ an action space $\mathcal{A}$, a transition dynamic function $P: \mathcal{S} \times \mathcal{A} \rightarrow \mathcal{P}(\mathcal{S})$, a reward function $r:\mathcal{S} \times \mathcal{A}\rightarrow \mathbb{R}$, an initial state distribution $\rho_0\in \mathcal{P}(\mathcal{S})$, and a discount factor $\gamma \in [0, 1)$.

% Offline RL aims to learn the optimal policy $\pi^*$ using only a fixed offline dataset $D_{\text{offline}}=\{(s_i,a_i,r_i,s_i')\}$, a collection of the state $s_i$, action $a_i$, reward $r_i$, and the next state $s_i'$, without any additional interaction with the environment. Offline datasets can be collected by an unknown behavior policy $\pi_\mu$, which can lead to instability in value function estimation when the learned policy attempts out-of-distribution (OOD) action outside the support of the dataset.

Offline RL aims to learn the optimal policy $\pi^*$ using only a fixed offline dataset $D_{\text{offline}}=\{(s_i,a_i,r_i,s_i')\}$, which is a collection of transitions typically gathered by an unknown behavior policy $\pi_\mu$ without any additional interaction with the environment. Offline datasets can lead to instability in value function estimation when the learned policy attempts out-of-distribution (OOD) action outside the support of the dataset.

In the setting of O2O RL, the agent leverages a policy $\pi_\text{offline}$ pre-trained with $D_{\text{offline}}$ as an initial policy for online fine-tuning in an online environment. During the fine-tuning phase, the agent interacts with the environment, collects new transition data $(s_i,a_i,r_i,s_i')$, and adds it to $D_\text{online}$. Both the offline dataset $D_\text{offline}$ and the newly collected online dataset $D_\text{online}$ are utilized to progressively improve the policy. The goal of O2O RL is to achieve higher final performance through online interaction while minimizing initial performance degradation due to OOD issues.
\section{Method}
This section details the methodology for the Adaptive Replay Buffer (ARB), designed to achieve optimal performance during the online fine-tuning phase of O2O RL.
\subsection{Motivation of ARB and On-policyness}
The idea of ARB comes from the fact that online fine-tuning is fundamentally an off-policy learning process with on-policy data collection. Previous works \citep{sutton1998reinforcement, schulman2015high} have argued that learning from on-policy data reduces learning variance and improves final performance. Building on this insight, we hypothesize that an effective data-mixing strategy should not only balance offline and online data but also prioritize the most valuable data within this mix: the transition that are most congruent with the current policy.

We define the on-policyness of a given transition $(s,a,r,s')$ for a policy $\pi_\theta$ is defined as the likelihood of that transition within an on-policy dataset, a dataset collected by the policy $\pi_\theta$:
\begin{equation}
\mathcal{O}(s,a,r,s';\pi_\theta)=d^{\pi_\theta}(s)\pi_\theta(a|s)
\end{equation}
where $\pi_\theta(a|s)$ is the likelihood of the current policy, and $d^{\pi_\theta}(s)$ represents the stationary state distribution under that policy. This metric allows us to precisely measure how relevant a past experience is to the agent's current behavior, forming the foundation of adaptive sampling strategy.

However, directly computing the full stationary distribution, $d^{\pi_\theta}(s)$, is generally intractable. Estimating $d^{\pi_\theta}(s)$ explicitly in continuous-state environments with unknown dynamics is highly complex and computationally prohibitive. Furthermore, relying too heavily on this term could inadvertently lead the agent to over-prioritize states it has already extensively explored, potentially inhibiting further exploration and trapping it in a local optimum. To address these challenges, we approximate $d^{\pi_\theta}(s)$ as a uniform distribution to focus solely on the action probability, as follows:
\begin{equation}
\tilde{\mathcal{O}}(s,a,r,s';\pi_\theta)=\pi_\theta(a|s)
\end{equation}
This value is not only straightforward to compute but also provides a direct measure of how aligned a given transition is with the agent's current behavior. By sampling data proportional to this on-policyness, our replay buffer effectively prioritizes experiences that are most relevant to the current policy's actions, ensuring a balance between leveraging valuable data for refinement and encouraging continued exploration without explicit state-space bias. Since the on-policyness of a transition is not relevant to its reward and the next state, we simply denote $\tilde{\mathcal{O}}(s,a,r,s';\pi_\theta)$ as $\tilde{\mathcal{O}}(s,a;\pi_\theta)$

\subsection{Adaptive Replay Buffer}

This section explains how to build an Adaptive Replay Buffer (ARB) by leveraging on-policyness. The ARB is designed to be lightweight, easy to implement, and learning-free.

% First, the on-policyness value for each transition within the dataset is calculated relative to the current policy. This is done by utilizing the policy's probability mass function for discrete action spaces and the probability density function for continuous action spaces.

First, the on-policyness value for each transition within the dataset is calculated relative to the current policy. This is done by utilizing the policy's probability mass function for discrete action spaces and the probability density function for continuous action spaces. For computational stability, ARB clamps the log-likelihood of the policy values between $\underline{p}$ and $\bar{p}$ and subtracts the maximum log-likelihood value among all transitions. The on-policyness value is then bounded within the range $\left( 0,1\right]$ while maintaining the original sampling proportions for all transitions:
\begin{equation}
\label{eq:onpolicyness}
\begin{aligned}
\tilde{\mathcal{O}}(s,a;\pi_\theta)&=\exp\left(\text{Clip}\left({\log \pi_{\theta} (a_t|s_t), \underline{p}, \bar{p}}\right) - p_{\max}\right),\\
p_{\max}&=\max_{d\in\mathcal{D}}\left({\text{Clip}\left(\log\pi_\theta(a_t|s_t), \underline{p}, \bar{p}\right)}\right)
\end{aligned}
\end{equation}
where $p_{\max}$ is the maximum value of the bounded log-probability value among all transitions $d\in\mathcal{D}$.

A naive approach would sample transitions from the replay buffer with a probability proportional to their individual on-policyness value. However, the probability value for a single transition can have high variance, which poses a risk of repeatedly sampling a small number of transitions. To improve training stability, the ARB calculates the on-policyness at the trajectory level. Each transition within a trajectory then uses that trajectory's on-policyness value for sampling. The on-policyness of a trajectory $\tau=(s_0,a_0,r_0,s_1,...,s_T)$ is defined as the geometric mean of the on-policyness values of its constituent transitions. This trajectory-level prioritization ensures a smoother and more stable training process by reducing the variance of sampling probabilities. The sampling weight of a transition $(s,a,r,s')$ contained in a trajectory $\tau=(s_0,a_0,r_0,...,s_T)$ in ARB is given by:
% \begin{equation}
% \label{eq:sampling_weight}
% \begin{aligned}
% \omega(s,a,r,s')&=\left(\prod_{t=0}^{T-1}{\tilde{\mathcal{O}}(s_t,a_t;\pi_\theta)}\right)^{\frac{1}{|\tau|}}\\
% &=\exp\left(\frac{1}{|\tau|}\sum_{t=0}^{T-1}\frac{p(s_t,a_t;\pi_\theta)}{\lambda}\right)\\
% \end{aligned}
% \end{equation}
\begin{equation}
\label{eq:sampling_weight}
\begin{aligned}
\omega(s,a,r,s')&=\left(\prod_{t=0}^{T-1}{\tilde{\mathcal{O}}(s_t,a_t;\pi_\theta)^{\frac{1}{\lambda}}}\right)^{\frac{1}{|\tau|}}\\
&=\exp\left(\frac{1}{|\tau|}\sum_{t=0}^{T-1}\frac{\log \tilde{\mathcal{O}}(s_t,a_t;\pi_\theta)}{\lambda}\right)\\
\end{aligned}
\end{equation}
where $\tau$ is a trajectory that contains the transition $(s,a,r,s')$ and $\lambda>0$ is a temperature parameter that controls the degree to which sampling is proportional to on-policyness. As the temperature $\lambda>0$ grows smaller, the sampling proportion is more dominated by the on-policyness.

\subsection{Offline-to-Online RL with ARB}
The overall training process with ARB involves two distinct phases: offline pre-training and online fine-tuning. The complete process is detailed in \Cref{alg:arb}.
\subsubsection{Offline Pre-training}
This initial phase focuses on leveraging a pre-collected, static dataset $\mathcal{D}_{\text{offline}}$. The goal here is to learn a robust starting policy and value function without any interaction with the environment. As shown in Algorithm 1, this phase proceeds for a predefined number of iterations $N_{\text{pretrain}}$. In each iteration, a minibatch of transitions $\mathcal{M}$ is uniformly sampled from the offline dataset. This data is then used to learn the policy and value function, following the objective of the base O2O RL algorithm. This pre-training process provides a stable and knowledgeable foundation, which serves as a crucial "warm start" for the subsequent online fine-tuning.

\subsubsection{Online Fine-tuning}
Following the pre-training phase, the agent transitions to online fine-tuning, where it interacts with the environment and adapts its policy. This phase begins by initializing an empty ARB, $\mathcal{B}$, and then loading the entire offline dataset $\mathcal{D}_{\text{offline}}$ into it. 

\begin{algorithm}[H]
\caption{O2O RL with Adaptive Replay Buffer}
\label{alg:arb}
\begin{algorithmic}[1]
\Require Base O2O RL algorithm $\mathcal{A}$, offline dataset $\mathcal{D}_{\text{offline}}$, environment $Env$
\Require Hyperparameters: temperature $\lambda$, priority clips $[\underline{p}, \bar{p}]$, buffer re-weight frequency $d_{\text{weight}}$, model update frequency $d_{\text{update}}$, number of steps per update $N_{\text{update}}$

\State Initialize policy $\pi_\theta$.

\vspace{0.5em}
% \State \textcolor{blue}{Offline Pre-training Phase}
\State Phase 1: Offline Pre-training 
\For{$k=1$ \textbf{to} $N_{\text{pretrain}}$}
    \State Sample $\mathcal{M} \sim \mathcal{D}_{\text{offline}}$ (uniformly).
    \State Update $\theta \leftarrow \mathcal{A}(\mathcal{M},\theta)$.
\EndFor

\vspace{0.5em}
% \State \textcolor{blue}{Online Fine-tuning Phase}
\State Phase 2: Online Fine-tuning
\State Initialize ARB $\mathcal{B}=\emptyset$.
\State $\mathcal{B}\leftarrow \mathcal{B}\cup \mathcal{D}_{\text{offline}}$.
\For{step $e=1$ \textbf{to} $N_{\text{interaction}}$}
    \State Collect $(s,a,r,s')$ with $a\sim \pi_\theta(\cdot|s)$ in $Env$.
    \State $\mathcal{B}\leftarrow \mathcal{B}\cup\{(s,a,r,s')\}$.
    \If{$e$ \% $d_{\text{weight}} = 0$}
        \For{$(s,a,r,s')$ in $\mathcal{B}$}
            \State Calculate $\omega(s,a,r,s')$ with (\ref{eq:sampling_weight}).
        \EndFor
    \EndIf
    \If {$e$ \% $d_{\text{update}} = 0$}
        \For{$k=1$ \textbf{to} $N_{\text{update}}$}
            \State Sample $\mathcal{M}\sim\mathcal{B}$ (weighted sampling).
            \State Update $\theta \leftarrow \mathcal{A}(\mathcal{M},\theta)$.
        \EndFor
    \EndIf
\EndFor
\end{algorithmic}
\end{algorithm}

The agent then starts to interact with the environment for $N_{\text{interaction}}$ steps. In each step, the agent executes the current policy to collect a new transition and adds the transition to $\mathcal{B}$. At each $d_{\text{weight}}$ steps, the agent recalculates the sampling weights for all transitions in the buffer. This re-weighting is based on the on-policyness metric defined in \cref{eq:sampling_weight}, ensuring that the sampling priorities reflect the agent's most current behavior. At each $d_{\text{update}}$ steps, the agent samples a minibatch $\mathcal{M}$ from $\mathcal{B}$ and learns the policy and value function for $N_{\text{update}}$ steps.

By using this adaptive, weighted sampling approach, the ARB ensures that the policy is trained on a dynamic mixture of both old offline data and new online data, with a strong emphasis on the experiences most relevant to its current behavior. This allows the agent to effectively balance stability from the pre-training phase with rapid adaptation during fine-tuning, mitigating the risks of catastrophic forgetting and accelerating performance improvement.

\section{Experiment}

This section presents a series of experiments designed to validate the efficacy of the proposed Adaptive Replay Buffer (ARB) and to analyze its core mechanisms. The experiments are structured to address the following three research questions:

1) Does ARB consistently outperform existing replay buffer strategies for Offline-to-Online Reinforcement Learning (O2O RL)?

2) Does ARB effectively and adaptively manage data composition, particularly by prioritizing online data when the offline data is of low quality and retaining offline data when it is highly valuable?

3) How do ARB's key hyperparameters and training settings, such as temperature $\lambda$, influence its performance and adaptive behavior?

\subsection{Experimental Setup}

We conduct our experiments on the three D4RL locomotion environments (\texttt{hopper, walker2d, halfcheetah}) with four different levels of offline datasets (\texttt{random}, \texttt{medium-replay}, \texttt{medium}, \texttt{medium-expert}) and six Antmaze environments. To demonstrate ARB's versatility, we integrate ARB into three widely applied O2O RL algorithms: Cal-QL \citep{nakamoto2023cal}, PEX \citep{zhang2023policy}, and FamO2O \citep{wang2023train}.

We compare ARB against a set of representative replay buffer strategies:
\begin{itemize}
    \item Naive: The simplest approach, where the replay buffer contains all offline data, and newly collected online data is appended. Samples are drawn uniformly from the entire buffer.
    \item Parallel: A fixed-ratio approach that samples 50\% of its batch from the offline dataset and 50\% from the online dataset.
    \item Top-N: An offline data-filtering method that retains only the top-N best-performing trajectories from the offline dataset (based on cumulative reward), discarding the rest. All online data is then added, and sampling is performed uniformly.
    \item Balanced Experience Replay Buffer (BERB) \citep{lee2022offline}: We implement the prioritized replay mechanism as a strong, data-aware baseline for comparison.
\end{itemize}

All methods are first pre-trained for 1M steps on the offline dataset and then fine-tuned online for an additional 1M environment steps. We evaluate performance by averaging the normalized scores over four random seeds and report the mean.

\subsection{O2O RL Performance Comparison}

\begin{table*}[ht!]
\caption{Average normalized returns of O2O RL algorithms across different replay buffer strategies on D4RL benchmarks. Results are grouped by the quality of offline dataset. (r: random, mr: medium-replay, m: medium, me: medium-expert)}
\label{tab:algorithm_by_method}
\centering
\begin{tabular}{ll|c|c|c|c|c}
\toprule
\textbf{Environment Set} & \textbf{Algorithm} & \textbf{Naive} & \textbf{Parallel} & \textbf{Top-N} & \textbf{BERB} & \textbf{ARB (Ours)} \\
\midrule
\multirow{4}{*}{\textbf{Locomotion (r)}} & \textbf{Cal-QL} & 10.6 & 11.6 & 10.9 & 11.2 & \textbf{17.9} \\
& \textbf{PEX} & 15.4 & 15.3 & 13.4 & 14.9 & \textbf{24.9} \\
& \textbf{FamO2O} & 30.5 & 40.9 & 26.3 & 35.9 & \textbf{53.1} \\
\cline{2-7}
& \textbf{Average} & 18.8 & 22.6 & 16.9 & 20.7 & \textbf{32.0} \\
\midrule
\multirow{4}{*}{\textbf{Locomotion (mr)}} & \textbf{Cal-QL} & 81.9 & 82.9 & 84.5 & 83.0 & \textbf{86.0} \\
& \textbf{PEX} & 58.4 & 57.9 & 76.2 & 61.7 & \textbf{82.9} \\
& \textbf{FamO2O} & 86.0 & 84.5 & 81.5 & 86.2 & \textbf{90.8} \\
\cline{2-7}
& \textbf{Average} & 75.4 & 75.1 & 80.7 & 77.0 & \textbf{86.6} \\
\midrule
\multirow{4}{*}{\textbf{Locomotion (m)}} & \textbf{Cal-QL} & 69.0 & 81.7 & \textbf{86.1} & 80.0 & 86.0 \\
& \textbf{PEX} & 67.4 & 73.2 & 72.3 & 74.6 & \textbf{75.7} \\
& \textbf{FamO2O} & 74.5 & 92.0 & 80.8 & 86.2 & \textbf{98.3} \\
\cline{2-7}
& \textbf{Average} & 70.3 & 82.3 & 79.7 & 80.3 & \textbf{86.7} \\
\midrule
\multirow{4}{*}{\textbf{Locomotion (me)}} & \textbf{Cal-QL} & 106.1 & 106.5 & \textbf{107.6} & 106.7 & 106.9 \\
& \textbf{PEX} & 82.7 & 80.8 & 83.3 & 83.7 & \textbf{86.2} \\
& \textbf{FamO2O} & 103.7 & 105.9 & 100.3 & 105.5 & \textbf{107.9} \\
\cline{2-7}
& \textbf{Average} & 97.5 & 97.7 & 97.1 & 98.6 & \textbf{100.3} \\
\midrule
\multirow{4}{*}{\textbf{Antmaze}} & \textbf{Cal-QL} & 76.6 & 81.3 & 56.5 & 75.3 & \textbf{85.2} \\
& \textbf{PEX} & 83.4 & 87.4 & 87.1 & 86.5 & \textbf{91.0} \\
& \textbf{FamO2O} & 75.8 & 85.8 & 83.2 & 83.3 & \textbf{88.4} \\
\cline{2-7}
& \textbf{Average} & 78.6 & 84.8 & 75.6 & 81.7 & \textbf{88.2} \\
\bottomrule
\end{tabular}
\end{table*}

This experiment evaluates the effectiveness of ARB to the performance of O2O RL algorithms against several established replay buffer strategies across various D4RL benchmarks, summarized in \Cref{tab:algorithm_by_method}. The result demonstrates that ARB consistently and significantly outperforms all baselines, regardless of the base O2O RL algorithm or environment difficulty.

The performance gains of ARB are particularly pronounced in environments with relatively low quality offline data, such as the Locomotion \texttt{random} and \texttt{medium-replay} datasets. This stark difference can be attributed to our method's ability to dynamically prioritize high-on-policyness data, effectively and quickly filtering out unhelpful, low-reward transitions that would otherwise hinder learning. For high-quality datasets like Locomotion \texttt{medium} and \texttt{medium-expert}, where the offline data already contains mid- and expert-level trajectories, ARB continues to hold a performance advantage, albeit with a smaller margin. This demonstrates that even when the offline data is of high quality, ARB's dynamic sampling provides a consistent edge, allowing the agent to fine-tune more effectively. The Antmaze environments also follow this trend, with ARB consistently achieving the best scores across different algorithms.

These results validate our core hypothesis: a dynamic replay buffer that prioritizes data based on its relevance to the current policy is a more robust and effective strategy than static mixing ratios or simple filtering. ARB's adaptive nature proves most impactful in challenging data conditions, but its benefits are consistent across all environments, confirming its versatility as a plug-in solution.

\subsection{Online-Data Ratio Analysis}

\begin{figure}[ht]
    \centering
    \includegraphics[width=1.0\columnwidth]{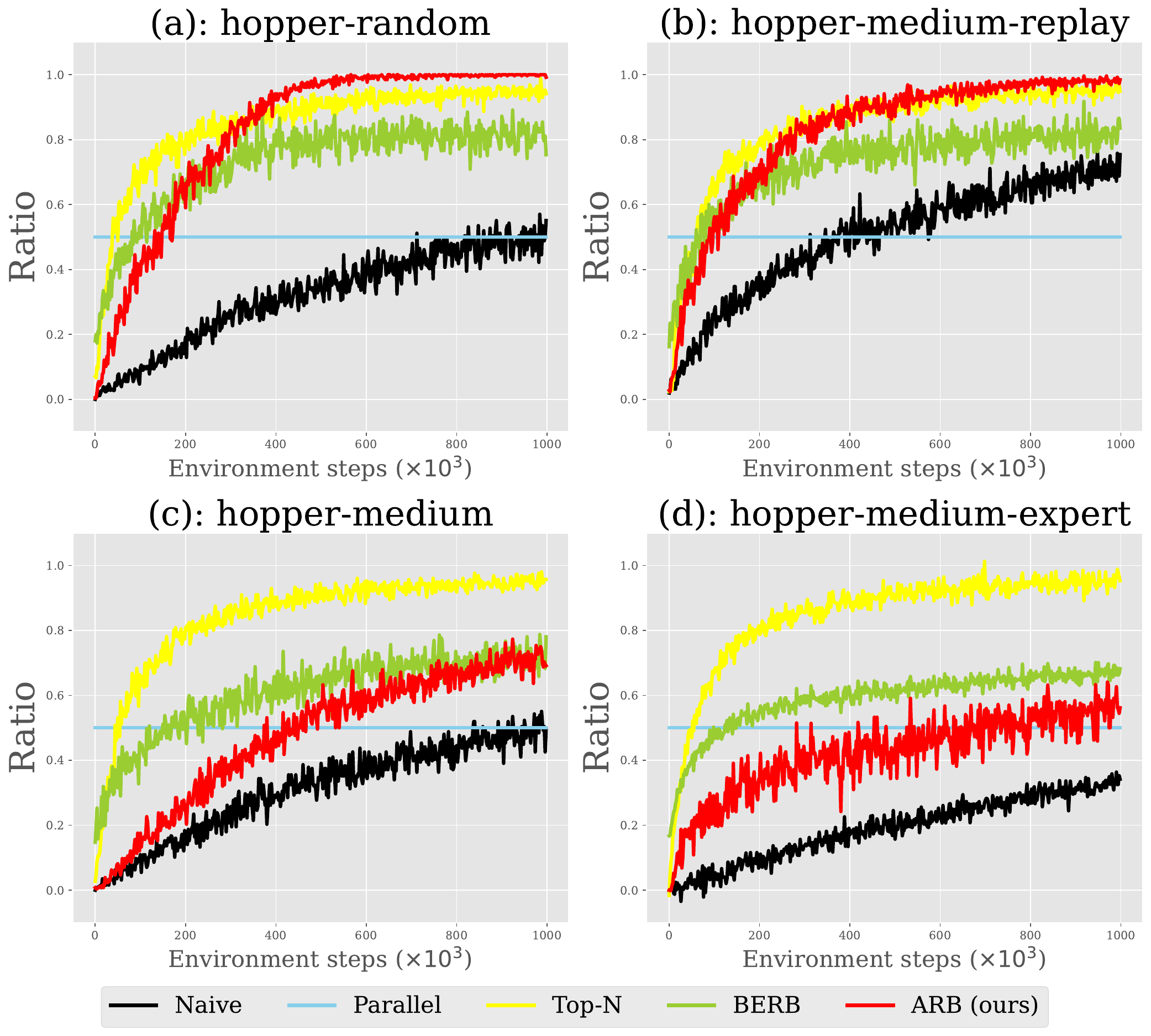}
    \caption{Online data ratio of the minibatch over environment steps for different \texttt{hopper} datasets with FamO2O}
    \label{fig:hopper_results}
\end{figure}

We analyze the online data sampling ratio of the minibatch across different D4RL datasets of varying quality. We use the \texttt{hopper} locomotion environments with four different dataset quality levels (\texttt{random, medium-replay, medium, medium-expert}) with FamO2O. Our hypothesis is that ARB will be more aggressive in prioritizing online data when the offline data quality is low, as the offline data is less valuable for improving the policy.

\Cref{fig:hopper_results} illustrates the change in the online data sampling ratio over 1M environment steps, with a grey line indicating the overall ratio of online data in the entire dataset. As shown in \Cref{fig:hopper_results}, ARB's online data ratio curve is consistently higher than the grey line, indicating that ARB is inherently biased towards sampling newer, online data. This can be attributed to the fact that online data is, by nature, more likely to be on-policy and therefore more relevant to the current policy's behavior.

Furthermore, a critical finding emerges when the offline dataset's average reward is low. In these cases, ARB's online data ratio curve rises sharply, a behavior not observed in other methods. This effect provides direct evidence of ARB's adaptive prioritization mechanism. By performing on-the-fly prioritized sampling, ARB is able to quickly identify and discard offline data that is too far from the current policy's distribution and thus unhelpful for performance improvement. This allows the agent to rapidly shift its focus to the more valuable online experiences, accelerating the fine-tuning process.

\subsection{Ablation Studies}

\begin{figure}[ht]
    \centering
    \includegraphics[width=1.0\columnwidth]{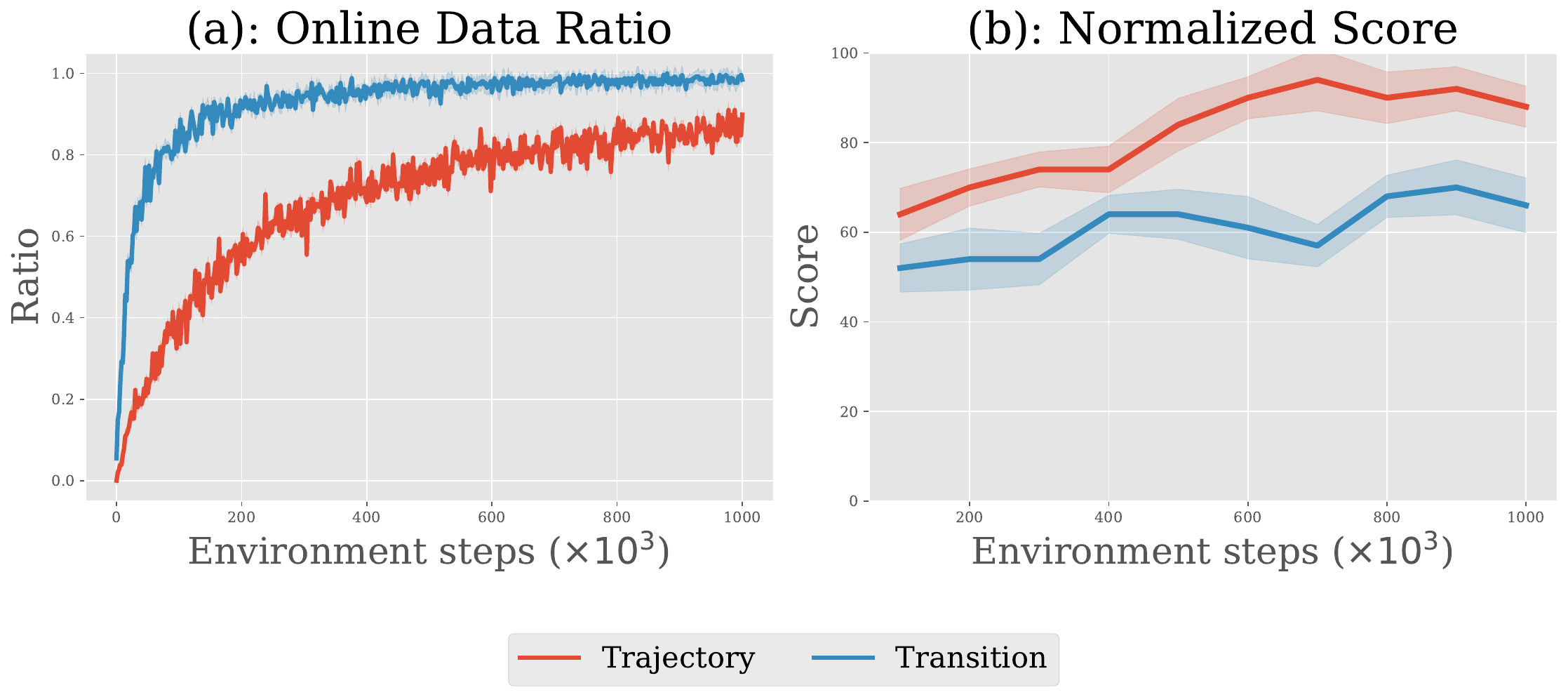}
    \caption{Normalized scores and online data ratios presented for the antmaze-large-play environment with Cal-QL, with respect to trajector}
    \label{fig:ablation_traj}
\end{figure}

We conduct a series of ablation studies to analyze the impact of our core design choices on the performance and behavior of ARB. For these studies, we focus on isolating the effects of key components, which we evaluate in a single environment.

\subsubsection{Trajectory vs Transition ARB}

\begin{figure}[ht]
    \centering
    \includegraphics[width=1.0\columnwidth]{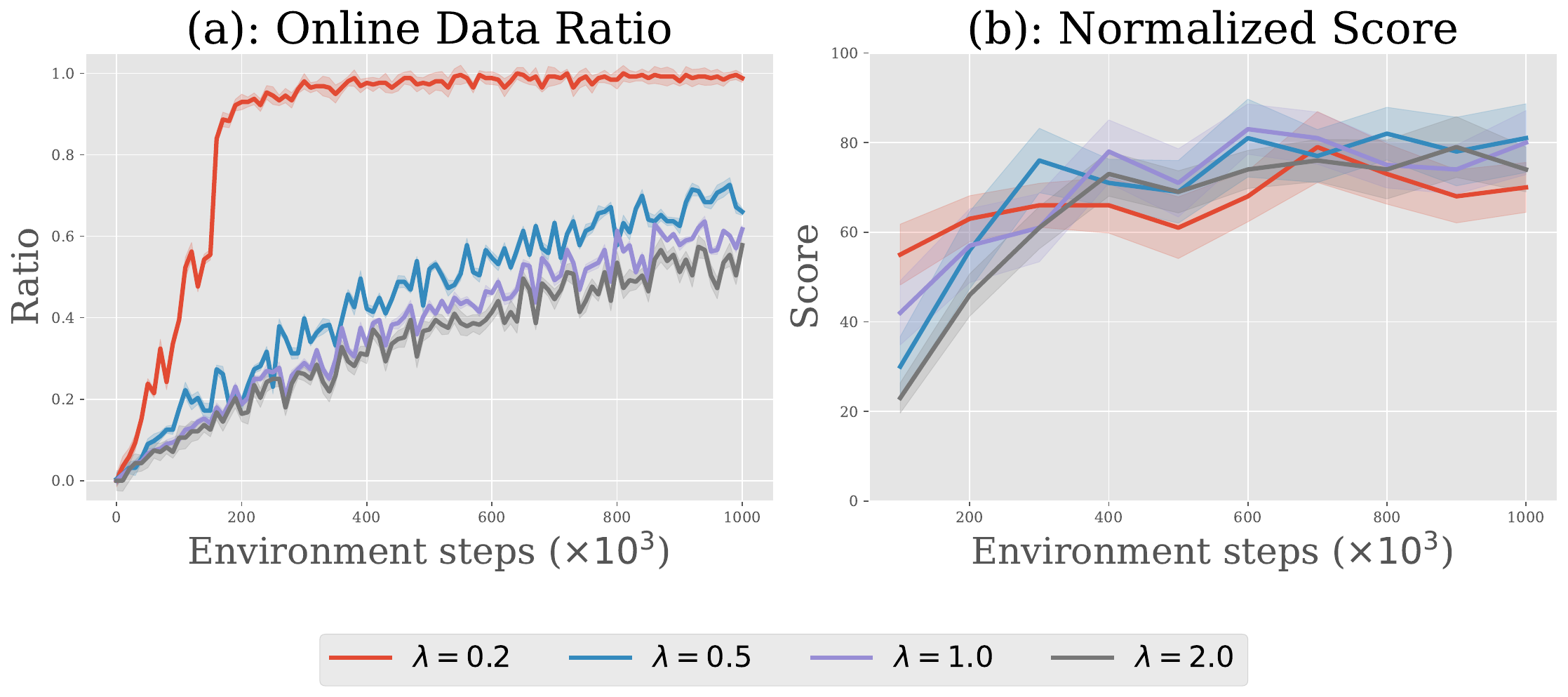}
    \caption{Online data ratio and normalized score presented for the \texttt{antmaze-large-diverse} environment with PEX, with respect to various temperature values}
    \label{fig:ablation_lambda}
\end{figure}

To further validate our design choices, we conducted an ablation study to assess the importance of calculating on-policyness at the trajectory level. It prevents specific data points from being over-sampled, which can compromise training stability. We performed this ablation study on the \texttt{antmaze-large-play-v2} environment using the Cal-QL algorithm. The results are presented in \Cref{fig:ablation_traj}.

Transition-based sampling shows a higher online data ratio because the exponential of the geometric mean is less than the average of the exponential values. \Cref{fig:ablation_traj} shows that the trajectory-based method achieves a significantly higher final normalized score compared to the transition-based method, suggesting that our hypothesis that aggregating on-policyness at the trajectory level is crucial for mitigating high-variance sampling and ensuring robust training, leading to superior final performance. By distributing the sampling weight across an entire sequence, ARB effectively prevents the policy from over-fitting to a few highly relevant but potentially noisy transitions.

\subsubsection{Temperature Parameter \texorpdfstring{$\lambda$}{lambda}}

The hyperparameter $\lambda$ in ARB's sampling priority equation controls the weighting of the on-policyness metric. We test four different values of $\lambda$ to discover its effect on the online data ratio and the O2O RL performance.  We perform these experiments on the \texttt{antmaze-large-diverse-v2} using the PEX algorithm.

As shown in \Cref{fig:ablation_lambda}, the online data ratio increases at a faster rate as the temperature value decreases. This directly confirms that $\lambda$ is a crucial hyperparameter for controlling the effect of on-policyness within our ARB. A lower temperature value makes the sampling distribution more sensitive to high on-policyness scores, thereby prioritizing recent, on-policy data more aggressively.

The normalized scores also follow a similar trend, showing that adjusting the temperature value allows for a clear trade-off between stability and adaptation. By carefully tuning this hyperparameter, researchers can find an optimal balance to accelerate online fine-tuning and improve final performance.

\subsection{Clipping Bound $\underline{p}$ and $\bar{p}$}

\begin{table}[ht]
\centering
\caption{Sensitivity analysis of the clipping bounds on \texttt{antmaze-large-diverse-v2}. Default values are $\underline{p}=-12.0$ and $\overline{p}=7.0$.}
\label{tab:sensitivity_bounds}
\small
\begin{tabular}{cc|cc}
\toprule
$\underline{p}$ & \textbf{Score} & $\overline{p}$ & \textbf{Score} \\
\midrule
$-16.0$ & $75.75 \pm 2.16$ & $2.0$  & $77.50 \pm 7.27$ \\
$-14.0$ & $78.75 \pm 1.71$ & $5.0$  & $77.00 \pm 5.35$ \\
$-12.0$ & $82.00 \pm 2.94$ & $7.0$  & $82.00 \pm 2.94$ \\
$-9.0$  & $82.75 \pm 3.40$ & $9.0$  & $70.25 \pm 4.11$ \\
$-7.0$  & $77.50 \pm 2.52$ & $12.0$ & $29.50 \pm 34.41$ \\
\bottomrule
\end{tabular}
\end{table}

We conduct a sensitivity analysis on the clipping bounds, $\underline{p}$ and $\overline{p}$, which are used to stabilize the on-policyness calculation by bounding the log-likelihood values. We evaluate the performance on the \texttt{antmaze-large-diverse-v2} environment using the PEX algorithm. The tested values for the clipping bounds are selected as approximations of $\ln 10^n$. The normalized scores are averaged over four random seeds, and we report the mean and standard deviation.

As shown in \Cref{tab:sensitivity_bounds}, the performance of ARB is generally robust and shows comparable results across different values of the lower bound $\underline{p}$ and moderate values of the upper bound $\overline{p}$. However, the performance degrades significantly when the upper bound $\overline{p}$ is set to excessively high values (e.g., $\overline{p}=9.0$ and $\overline{p}=12.0$). This indicates that an excessively high $\overline{p}$ can lead to the over-sampling of a small subset of data. Consequently, this results in the limited utilization of the available dataset, which hinders the agent's ability to learn a generalized policy during the fine-tuning phase.
\section{Conclusion}

In this paper, we introduced the Adaptive Replay Buffer (ARB) to solve the critical data-mixing dilemma in Offline-to-Online Reinforcement Learning (O2O RL). By dynamically prioritizing data based on a lightweight on-policyness metric, ARB effectively balances the stability of offline knowledge with the adaptability of new online experiences. Our extensive experiments on D4RL benchmarks showed that ARB consistently outperformed all baselines, particularly on low-quality datasets where its adaptive sampling mechanism proved most advantageous. This work demonstrates the power of a behavior-aware replay buffer design and offers a versatile solution for boosting the performance of modern O2O RL systems.
\clearpage

\bibliography{reference}

\section*{Checklist}

\begin{enumerate}

  \item For all models and algorithms presented, check if you include:
  \begin{enumerate}
    \item A clear description of the mathematical setting, assumptions, algorithm, and/or model. [Yes]
    \item An analysis of the properties and complexity (time, space, sample size) of any algorithm. [Yes]
    \item (Optional) Anonymized source code, with specification of all dependencies, including external libraries. [Yes]
  \end{enumerate}

  \item For any theoretical claim, check if you include:
  \begin{enumerate}
    \item Statements of the full set of assumptions of all theoretical results. [Not Applicable]
    \item Complete proofs of all theoretical results. [Not Applicable]
    \item Clear explanations of any assumptions. [Yes]     
  \end{enumerate}

  \item For all figures and tables that present empirical results, check if you include:
  \begin{enumerate}
    \item The code, data, and instructions needed to reproduce the main experimental results (either in the supplemental material or as a URL). [Yes]
    \item All the training details (e.g., data splits, hyperparameters, how they were chosen). [Yes]
    \item A clear definition of the specific measure or statistics and error bars (e.g., with respect to the random seed after running experiments multiple times). [Yes]
    \item A description of the computing infrastructure used. (e.g., type of GPUs, internal cluster, or cloud provider). [Yes]
  \end{enumerate}

  \item If you are using existing assets (e.g., code, data, models) or curating/releasing new assets, check if you include:
  \begin{enumerate}
    \item Citations of the creator If your work uses existing assets. [Yes]
    \item The license information of the assets, if applicable. [Yes]
    \item New assets either in the supplemental material or as a URL, if applicable. [Yes]
    \item Information about consent from data providers/curators. [Yes]
    \item Discussion of sensible content if applicable, e.g., personally identifiable information or offensive content. [Not Applicable]
  \end{enumerate}

  \item If you used crowdsourcing or conducted research with human subjects, check if you include:
  \begin{enumerate}
    \item The full text of instructions given to participants and screenshots. [Not Applicable]
    \item Descriptions of potential participant risks, with links to Institutional Review Board (IRB) approvals if applicable. [Not Applicable]
    \item The estimated hourly wage paid to participants and the total amount spent on participant compensation. [Not Applicable]
  \end{enumerate}

\end{enumerate}

%\section{Implementation Details}
%\subsection{Offline to Online RL Algorithms}
%We follow codebase. 
%\subsection{ARB}

% Appendix는 분리해서 제출해야 하므로 지워야함

\clearpage
\appendix
\thispagestyle{empty}

% Supplementary material: To improve readability, you must use a single-column format for the supplementary material.
\onecolumn
\section{IMPLEMENTATION DETAIL}

\subsection{O2O RL Algorithm Implementation}

For our experiments, we used a Clean Offline RL \citep{tarasov2022corl} codebase for the implementation of Cal-QL \citep{nakamoto2023cal}. PEX \citep{zhang2023policy} based on IQL, and FamO2O \citep{wang2023train} based on IQL were implemented based on their official codebases, which we adapted to be compatible with the same Clean Offline RL framework. The specific hyperparameter values used for each of these algorithms are detailed in \Cref{tab:hyper_calql}, \Cref{tab:hyper_pex}, and \Cref{tab:hyper_famo2o}.

For training, to ensure a fair comparison, we pretrain 1M steps for each algorithm and do 1M environment steps. At each $d_{\text{update}}=1000$ environment steps, we fine-tuned our model $N_{\text{update}}=1000$ steps. We sample 100 trajectories to measure the normalized score of the trained policy and use the average score. Our experiments were conducted on four machines with an AMD Ryzen Threadripper 2990WX 32-Core Processor CPU and an NVIDIA RTX 3090 GPU, using a software environment of Python 3.9, CUDA 11.3, and PyTorch 1.11.

\begin{table}[ht!]
\caption{Hyperparameters for Cal-QL}
\label{tab:hyper_calql}
\centering
\begin{tabular}{lc}
\toprule
\textbf{Hyperparameter} & \textbf{Locomotion} (\textbf{Antmaze}) \\
\midrule
Batch size & 256 \\
Discount factor & 0.99 \\
Regularizer Coefficient $\alpha$ & 10.0 (5.0) \\
Q hidden dims & 256 \\
Q hidden layers & 5 (3) \\
Q learning rate & $3\times10^{-4}$ \\
Policy hidden dims & 256 \\
Policy hidden layers & 3 \\
Policy learning rate & $3\times10^{-4}$ \\
target action gap & -1.0 (0.8) \\
soft target update rate & 0.005 \\
\# of CQL actions $k$ & 10 \\
\bottomrule
\end{tabular}
\end{table}

\begin{table}[ht!]
\caption{Hyperparameters for PEX based on IQL}
\label{tab:hyper_pex}
\centering
\begin{tabular}{lc}
\toprule
\textbf{Hyperparameter} & \textbf{Locomotion} (\textbf{Antmaze}) \\
\midrule
Batch size & 256 \\
Discount factor & 0.99 \\
PEX temperature $\beta_{\text{PEX}}$ & 3.0 (10.0) \\
IQL temperature $\beta$ & 3.0 (10.0) \\
Expectile $\tau$ & 0.7 (0.9) \\
Q hidden dims & 256 \\
Q hidden layers & 2 \\
Q learning rate & $3\times10^{-4}$ \\
V hidden dims & 256 \\
V hidden layers & 2 \\
V learning rate & $3\times10^{-4}$ \\
Policy hidden dims & 256 \\
Policy hidden layers & 2 \\
Policy learning rate & $3\times10^{-4}$ \\
soft target update rate & 0.005 \\
\bottomrule
\end{tabular}
\end{table}

\begin{table}[ht!]
\caption{Hyperparameters for FamO2O based on IQL}
\label{tab:hyper_famo2o}
\centering
\begin{tabular}{lc}
\toprule
\textbf{Hyperparameter} & \textbf{Locomotion} (\textbf{Antmaze}) \\
\midrule
Batch size & 256 \\
Discount factor & 0.99 \\
Balance coefficient min & 1.0 (8.0) \\
Balance coefficient min & 5.0 (14.0) \\
IQL temperature $\beta$ & 3.0 (10.0) \\
Expectile $\tau$ & 0.7 (0.9) \\
Q hidden dims & 256 \\
Q hidden layers & 2 \\
Q learning rate & $3\times10^{-4}$ \\
V hidden dims & 256 \\
V hidden layers & 2 \\
V learning rate & $3\times10^{-4}$ \\
Universal model hidden dims & 256 \\
Universal model hidden layers & 2 \\
Universal model learning rate & $3\times10^{-4}$ \\
Balance model hidden dims & 256 \\
Balance model hidden layers & 2 \\
Balance model learning rate & $3\times10^{-4}$ \\
balance coefficient dimension $l_b$ & 6 \\
soft target update rate & 0.005 \\
\bottomrule
\end{tabular}
\end{table}

\subsection{Baseline Replay Buffer Implementation}

We implemented the following replay  buffer baselines:
\begin{itemize}
    \item Naive: Samples were drawn uniformly from a replay buffer containing all offline data and newly collected online data.
    \item Parallel: A fixed-ratio approach where minibatches were composed of 50\% offline data and 50\% online data.
    \item Top-N: This method retained the top 50,000 transitions based on the best sum of rewards from the offline dataset. New online data was added to this buffer, and sampling was performed uniformly.
    \item Balanced Experience Replay Buffer: Implemented using the official codebase with specified hyperparameters \citep{lee2022offline} and adapted to our Clean offline RL framework.

\end{itemize}

\subsection{ARB Implementation Detail}

Our ARB was integrated into three base algorithms: Cal-QL, PEX, and FamO2O. Since all these policies are parameterized by a Gaussian distribution, the log-likelihood of a transition was calculated using the Probability Density Function of a standard normal distribution. This log-likelihood was then normalized by the number of action dimensions. We use the temperature $\lambda=5.0$ for Cal-QL and $\lambda=0.5$ for PEX and FamO2O. At each $d_{\text{weight}}=1000$ environment steps, we recalculate the sampling weight. For all environments and algorithms, we set $\underline{p}=-12.0$  and $\bar{p}=7.0$.

\section{DETAILED EXPERIMENT RESULTS}

\Cref{tab:d4rl_calql}, \Cref{tab:d4rl_pex} and \Cref{tab:d4rl_famo2o} show the normalized average returns on various D4RL environemnts with Cal-QL, PEX, FamO2O and IQL, respectively.

\begin{table*}[ht!]
\caption{Normalized average returns on various D4RL environments with Cal-QL.}
\label{tab:d4rl_calql}
\centering
\resizebox{\linewidth}{!}{
\begin{tabular}{l|c|c|c|c|c}
\toprule
\textbf{Environment} & \textbf{Naive} & \textbf{Parallel} & \textbf{Top-N} & \textbf{BERB} & \textbf{ARB(Ours)} \\
\midrule
\multicolumn{6}{l}{\textit{Locomotion Random}} \\
\midrule
halfcheetah-random-v2 & 22.77 $\pm$ 0.58 & 20.16 $\pm$ 0.60 & 17.59 $\pm$ 0.44 & 20.16 $\pm$ 0.57 & 24.25 $\pm$ 0.38 \\
hopper-random-v2 & 8.17 $\pm$ 0.24 & 7.22 $\pm$ 0.20 & 7.49 $\pm$ 0.35 & 7.47 $\pm$ 0.17 & 8.55 $\pm$ 0.25 \\
walker2d-random-v2 & 0.79 $\pm$ 0.13 & 7.30 $\pm$ 0.22 & 7.47 $\pm$ 0.28 & 6.03 $\pm$ 0.18 & 20.76 $\pm$ 0.31 \\
\midrule
\textbf{Average} & 10.58 $\pm$ 0.32 & 11.56 $\pm$ 0.34 & 10.85 $\pm$ 0.36 & 11.22 $\pm$ 0.31 & 17.85 $\pm$ 0.31 \\
\midrule
\multicolumn{6}{l}{\textit{Locomotion Medium Replay}} \\
\midrule
halfcheetah-medium-replay-v2 & 55.51 $\pm$ 0.49 & 58.42 $\pm$ 0.25 & 57.52 $\pm$ 0.34 & 57.67 $\pm$ 0.32 & 58.78 $\pm$ 2.13 \\
hopper-medium-replay-v2 & 101.09 $\pm$ 0.25 & 97.82 $\pm$ 0.37 & 100.55 $\pm$ 0.30 & 99.02 $\pm$ 0.31 & 102.54 $\pm$ 0.29 \\
walker2d-medium-replay-v2 & 89.14 $\pm$ 0.59 & 92.46 $\pm$ 0.29 & 95.51 $\pm$ 0.31 & 92.41 $\pm$ 0.10 & 96.66 $\pm$ 0.47 \\
\midrule
\textbf{Average} & 81.91 $\pm$ 0.44 & 82.90 $\pm$ 0.30 & 84.53 $\pm$ 0.32 & 83.03 $\pm$ 0.24 & 85.99 $\pm$ 0.96 \\
\midrule
\multicolumn{6}{l}{\textit{Locomotion Medium}} \\
\midrule
halfcheetah-medium-v2 & 58.07 $\pm$ 0.43 & 72.49 $\pm$ 0.32 & 77.42 $\pm$ 0.37 & 70.58 $\pm$ 0.09 & 75.65 $\pm$ 3.25 \\
hopper-medium-v2 & 66.83 $\pm$ 1.09 & 89.09 $\pm$ 0.44 & 95.66 $\pm$ 0.42 & 85.95 $\pm$ 0.08 & 96.69 $\pm$ 0.47 \\
walker2d-medium-v2 & 82.17 $\pm$ 0.55 & 83.56 $\pm$ 0.29 & 85.19 $\pm$ 0.33 & 83.55 $\pm$ 0.05 & 85.88 $\pm$ 2.87 \\
\midrule
\textbf{Average} & 69.02 $\pm$ 0.69 & 81.71 $\pm$ 0.35 & 86.09 $\pm$ 0.37 & 80.03 $\pm$ 0.07 & 86.07 $\pm$ 2.20 \\
\midrule
\multicolumn{6}{l}{\textit{Locomotion Medium Expert}} \\
\midrule
halfcheetah-medium-expert-v2 & 96.65 $\pm$ 0.39 & 97.69 $\pm$ 0.33 & 99.02 $\pm$ 0.38 & 97.74 $\pm$ 0.32 & 97.68 $\pm$ 0.30 \\
hopper-medium-expert-v2 & 111.98 $\pm$ 0.35 & 111.77 $\pm$ 0.35 & 112.38 $\pm$ 0.32 & 111.91 $\pm$ 0.22 & 112.21 $\pm$ 0.29 \\
walker2d-medium-expert-v2 & 109.62 $\pm$ 0.36 & 110.21 $\pm$ 0.18 & 111.39 $\pm$ 0.30 & 110.32 $\pm$ 0.09 & 110.75 $\pm$ 0.17 \\
\midrule
\textbf{Average} & 106.08 $\pm$ 0.37 & 106.56 $\pm$ 0.29 & 107.59 $\pm$ 0.33 & 106.66 $\pm$ 0.21 & 106.88 $\pm$ 0.25 \\
\midrule
\multicolumn{6}{l}{\textit{Random}} \\
\midrule
antmaze-umaze-v2 & 28.75 $\pm$ 14.15 & 49.25 $\pm$ 17.51 & 36.75 $\pm$ 21.01 & 42.50 $\pm$ 16.96 & 62.50 $\pm$ 10.38 \\
antmaze-umaze-diverse-v2 & 97.00 $\pm$ 1.73 & 98.75 $\pm$ 1.26 & 25.00 $\pm$ 50.00 & 83.75 $\pm$ 8.94 & 99.25 $\pm$ 1.26 \\
antmaze-medium-play-v2 & 95.50 $\pm$ 2.75 & 94.75 $\pm$ 2.75 & 94.75 $\pm$ 3.86 & 94.75 $\pm$ 2.06 & 94.75 $\pm$ 5.79 \\
antmaze-medium-diverse-v2 & 92.75 $\pm$ 4.57 & 92.75 $\pm$ 4.50 & 91.50 $\pm$ 6.95 & 92.50 $\pm$ 3.00 & 94.25 $\pm$ 3.10 \\
antmaze-large-play-v2 & 80.25 $\pm$ 4.11 & 82.00 $\pm$ 5.10 & 61.50 $\pm$ 42.22 & 77.50 $\pm$ 9.60 & 83.25 $\pm$ 6.80 \\
antmaze-large-diverse-v2 & 65.50 $\pm$ 15.52 & 70.00 $\pm$ 9.07 & 29.50 $\pm$ 33.50 & 61.00 $\pm$ 9.17 & 77.00 $\pm$ 5.66 \\
\midrule
\textbf{Average} & 76.96 $\pm$ 7.14 & 81.29 $\pm$ 6.54 & 56.5 $\pm$ 28.59 & 75.33 $\pm$ 8.13 & 85.17 $\pm$ 5.43 \\
\bottomrule
\end{tabular}
}
\end{table*}

\begin{table*}[ht!]
\caption{Normalized average returns on various D4RL environments with PEX.}
\label{tab:d4rl_pex}
\centering
\resizebox{\linewidth}{!}{
\begin{tabular}{l|c|c|c|c|c}
\toprule
\textbf{Environment} & \textbf{Naive} & \textbf{Parallel} & \textbf{Top-N} & \textbf{BERB} & \textbf{ARB(Ours)} \\
\midrule
\multicolumn{6}{l}{\textit{Locomotion Random}} \\
\midrule
halfcheetah-random-v2 & 25.18 $\pm$ 9.49 & 22.23 $\pm$ 10.43 & 21.30 $\pm$ 14.85 & 22.63 $\pm$ 11.01 & 37.56 $\pm$ 12.76 \\
hopper-random-v2 & 12.31 $\pm$ 4.25 & 12.86 $\pm$ 2.79 & 14.77 $\pm$ 3.28 & 12.99 $\pm$ 1.44 & 20.53 $\pm$ 5.50 \\
walker2d-random-v2 & 8.59 $\pm$ 2.41 & 10.67 $\pm$ 1.51 & 4.84 $\pm$ 1.19 & 9.09 $\pm$ 1.23 & 16.50 $\pm$ 3.64 \\
\midrule
\textbf{Average} & 15.36 $\pm$ 4.88 & 15.25 $\pm$ 4.91 & 13.64 $\pm$ 6.44 & 14.90 $\pm$ 4.56 & 24.86 $\pm$ 7.30 \\
\midrule
\multicolumn{6}{l}{\textit{Locomotion Medium Replay}} \\
\midrule
halfcheetah-medium-replay-v2 & 50.66 $\pm$ 5.95 & 52.59 $\pm$ 8.35 & 55.24 $\pm$ 5.73 & 52.74 $\pm$ 6.92 & 63.97 $\pm$ 3.56 \\
hopper-medium-replay-v2 & 50.34 $\pm$ 3.98 & 50.32 $\pm$ 12.91 & 64.77 $\pm$ 8.24 & 53.21 $\pm$ 9.59 & 74.15 $\pm$ 6.24 \\
walker2d-medium-replay-v2 & 74.12 $\pm$ 3.43 & 70.80 $\pm$ 3.97 & 82.79 $\pm$ 3.13 & 79.01 $\pm$ 2.77 & 106.56 $\pm$ 4.60 \\
\midrule
\textbf{Average} & 58.37 $\pm$ 4.45 & 57.90 $\pm$ 8.41 & 67.60 $\pm$ 5.70 & 61.65 $\pm$ 6.43 & 81.56 $\pm$ 4.80 \\
\midrule
\multicolumn{6}{l}{\textit{Locomotion Medium}} \\
\midrule
halfcheetah-medium-v2 & 46.99 $\pm$ 5.29 & 65.60 $\pm$ 6.84 & 71.10 $\pm$ 10.44 & 63.98 $\pm$ 5.83 & 69.65 $\pm$ 5.09 \\
hopper-medium-v2 & 70.16 $\pm$ 4.11 & 70.28 $\pm$ 16.80 & 70.49 $\pm$ 7.46 & 74.45 $\pm$ 9.14 & 74.12 $\pm$ 2.59 \\
walker2d-medium-v2 & 83.03 $\pm$ 3.04 & 83.72 $\pm$ 4.40 & 82.79 $\pm$ 4.08 & 85.30 $\pm$ 2.60 & 83.33 $\pm$ 3.29 \\
\midrule
\textbf{Average} & 66.73 $\pm$ 4.15 & 73.20 $\pm$ 9.35 & 74.79 $\pm$ 7.33 & 74.58 $\pm$ 5.86 & 75.71 $\pm$ 3.66 \\
\midrule
\multicolumn{6}{l}{\textit{Locomotion Medium Expert}} \\
\midrule
halfcheetah-medium-expert-v2 & 78.88 $\pm$ 3.09 & 72.45 $\pm$ 4.03 & 86.50 $\pm$ 3.42 & 76.54 $\pm$ 3.31 & 84.60 $\pm$ 3.48 \\
hopper-medium-expert-v2 & 60.42 $\pm$ 1.35 & 55.41 $\pm$ 4.10 & 77.55 $\pm$ 3.62 & 61.34 $\pm$ 2.26 & 64.19 $\pm$ 2.94 \\
walker2d-medium-expert-v2 & 108.86 $\pm$ 2.46 & 114.46 $\pm$ 2.97 & 113.35 $\pm$ 2.43 & 113.12 $\pm$ 0.86 & 109.96 $\pm$ 1.50 \\
\midrule
\textbf{Average} & 82.72 $\pm$ 2.30 & 80.77 $\pm$ 3.70 & 92.47 $\pm$ 3.16 & 83.67 $\pm$ 2.14 & 86.25 $\pm$ 2.64 \\
\midrule
\multicolumn{6}{l}{\textit{Random}} \\
\midrule
antmaze-umaze-v2 & 97.25 $\pm$ 1.64 & 96.25 $\pm$ 2.05 & 94.50 $\pm$ 2.06 & 96.00 $\pm$ 1.58 & 97.00 $\pm$ 0.71 \\
antmaze-umaze-diverse-v2 & 69.75 $\pm$ 8.18 & 93.25 $\pm$ 4.19 & 94.50 $\pm$ 2.08 & 88.75 $\pm$ 3.10 & 93.75 $\pm$ 4.99 \\
antmaze-medium-play-v2 & 93.25 $\pm$ 3.20 & 94.25 $\pm$ 3.10 & 91.50 $\pm$ 2.52 & 93.50 $\pm$ 1.73 & 95.50 $\pm$ 2.65 \\
antmaze-medium-diverse-v2 & 88.00 $\pm$ 10.13 & 81.50 $\pm$ 26.41 & 90.00 $\pm$ 4.32 & 84.50 $\pm$ 17.75 & 95.00 $\pm$ 3.56 \\
antmaze-large-play-v2 & 77.00 $\pm$ 4.69 & 78.25 $\pm$ 3.86 & 76.25 $\pm$ 6.45 & 77.50 $\pm$ 2.52 & 82.75 $\pm$ 3.40 \\
antmaze-large-diverse-v2 & 75.00 $\pm$ 7.87 & 81.00 $\pm$ 2.16 & 75.75 $\pm$ 1.71 & 78.75 $\pm$ 2.75 & 82.00 $\pm$ 2.94 \\
\midrule
\textbf{Average} & 83.38 $\pm$ 8.12 & 87.42 $\pm$ 10.45 & 87.08 $\pm$ 8.65 & 86.58 $\pm$ 8.28 & 91.00 $\pm$ 6.30 \\
\bottomrule
\end{tabular}
}
\end{table*}

\begin{table*}[ht!]
\caption{Normalized average returns on various D4RL environments with FamO2O.}
\label{tab:d4rl_famo2o}
\centering
\resizebox{\linewidth}{!}{
\begin{tabular}{l|c|c|c|c|c}
\toprule
\textbf{Environment} & \textbf{Naive} & \textbf{Parallel} & \textbf{Top-N} & \textbf{BERB} & \textbf{ARB(Ours)} \\
\midrule
\multicolumn{6}{l}{\textit{Locomotion Random}} \\
\midrule
halfcheetah-random-v2 & 21.05 $\pm$ 1.48 & 20.73 $\pm$ 0.90 & 19.30 $\pm$ 1.13 & 20.30 $\pm$ 0.93 & 24.33 $\pm$ 1.49 \\
hopper-random-v2 & 8.53 $\pm$ 0.94 & 7.98 $\pm$ 0.96 & 7.38 $\pm$ 0.14 & 8.13 $\pm$ 0.77 & 8.43 $\pm$ 0.94 \\
walker2d-random-v2 & 1.76 $\pm$ 0.77 & 7.43 $\pm$ 0.57 & 7.55 $\pm$ 0.36 & 6.65 $\pm$ 0.49 & 8.80 $\pm$ 1.60 \\
\midrule
\textbf{Average} & 10.45 $\pm$ 0.64 & 12.05 $\pm$ 0.48 & 11.41 $\pm$ 0.40 & 11.69 $\pm$ 0.43 & 13.85 $\pm$ 0.79 \\
\midrule
\multicolumn{6}{l}{\textit{Locomotion Medium Replay}} \\
\midrule
halfcheetah-medium-replay-v2 & 56.45 $\pm$ 1.41 & 57.00 $\pm$ 1.05 & 57.25 $\pm$ 0.94 & 57.18 $\pm$ 0.58 & 59.80 $\pm$ 1.63 \\
hopper-medium-replay-v2 & 99.40 $\pm$ 2.05 & 98.70 $\pm$ 2.15 & 98.85 $\pm$ 1.94 & 98.70 $\pm$ 1.48 & 99.43 $\pm$ 2.05 \\
walker2d-medium-replay-v2 & 89.20 $\pm$ 0.17 & 92.50 $\pm$ 0.52 & 95.80 $\pm$ 0.54 & 94.55 $\pm$ 0.65 & 96.55 $\pm$ 0.65 \\
\midrule
\textbf{Average} & 81.68 $\pm$ 0.83 & 82.73 $\pm$ 0.82 & 83.97 $\pm$ 0.74 & 83.48 $\pm$ 0.57 & 85.26 $\pm$ 0.90 \\
\midrule
\multicolumn{6}{l}{\textit{Locomotion Medium}} \\
\midrule
halfcheetah-medium-v2 & 64.93 $\pm$ 10.37 & 76.50 $\pm$ 1.70 & 75.33 $\pm$ 1.91 & 75.35 $\pm$ 0.46 & 77.20 $\pm$ 2.00 \\
hopper-medium-v2 & 73.10 $\pm$ 15.69 & 88.38 $\pm$ 10.16 & 94.60 $\pm$ 1.20 & 86.83 $\pm$ 5.21 & 96.03 $\pm$ 1.83 \\
walker2d-medium-v2 & 82.53 $\pm$ 0.73 & 84.83 $\pm$ 0.80 & 84.85 $\pm$ 0.96 & 83.50 $\pm$ 0.00 & 86.55 $\pm$ 1.55 \\
\midrule
\textbf{Average} & 73.52 $\pm$ 6.27 & 83.24 $\pm$ 3.44 & 84.93 $\pm$ 0.82 & 81.89 $\pm$ 1.74 & 86.59 $\pm$ 1.04 \\
\midrule
\multicolumn{6}{l}{\textit{Locomotion Medium-Expert}} \\
\midrule
halfcheetah-medium-expert-v2 & 97.40 $\pm$ 0.58 & 98.05 $\pm$ 0.44 & 98.93 $\pm$ 0.49 & 98.05 $\pm$ 0.27 & 98.05 $\pm$ 0.44 \\
hopper-medium-expert-v2 & 111.45 $\pm$ 0.39 & 111.60 $\pm$ 0.29 & 112.50 $\pm$ 0.16 & 112.10 $\pm$ 0.09 & 112.23 $\pm$ 0.11 \\
walker2d-medium-expert-v2 & 109.83 $\pm$ 0.08 & 110.13 $\pm$ 0.04 & 111.45 $\pm$ 0.09 & 110.33 $\pm$ 0.00 & 110.85 $\pm$ 0.17 \\
\midrule
\textbf{Average} & 106.23 $\pm$ 0.23 & 106.59 $\pm$ 0.18 & 107.63 $\pm$ 0.17 & 106.83 $\pm$ 0.09 & 107.04 $\pm$ 0.16 \\
\midrule
\multicolumn{6}{l}{\textit{Antmaze}} \\
\midrule
antmaze-umaze-v2 & 94.25 $\pm$ 0.96 & 95.25 $\pm$ 1.26 & 92.75 $\pm$ 0.50 & 94.50 $\pm$ 1.00 & 95.75 $\pm$ 2.87 \\
antmaze-umaze-diverse-v2 & 62.25 $\pm$ 10.97 & 91.75 $\pm$ 2.75 & 93.50 $\pm$ 2.65 & 86.25 $\pm$ 1.26 & 92.25 $\pm$ 4.72 \\
antmaze-medium-play-v2 & 91.25 $\pm$ 0.96 & 93.00 $\pm$ 4.55 & 89.00 $\pm$ 2.71 & 92.00 $\pm$ 3.16 & 95.00 $\pm$ 1.63 \\
antmaze-medium-diverse-v2 & 78.00 $\pm$ 34.69 & 80.25 $\pm$ 25.54 & 84.75 $\pm$ 6.70 & 80.75 $\pm$ 21.20 & 91.25 $\pm$ 2.87 \\
antmaze-large-play-v2 & 63.75 $\pm$ 4.35 & 77.00 $\pm$ 6.16 & 66.25 $\pm$ 4.27 & 72.25 $\pm$ 4.11 & 77.25 $\pm$ 1.26 \\
antmaze-large-diverse-v2 & 65.00 $\pm$ 8.04 & 77.25 $\pm$ 7.18 & 73.00 $\pm$ 3.16 & 74.25 $\pm$ 5.62 & 78.75 $\pm$ 4.27 \\
\midrule
\textbf{Average} & 75.75 $\pm$ 19.05 & 85.75 $\pm$ 12.76 & 83.21 $\pm$ 10.96 & 83.33 $\pm$ 11.82 & 88.38 $\pm$ 8.17 \\
\bottomrule
\end{tabular}
}
\end{table*}

\section{ADDITIONAL EXPERIMENT ON PURE OFFLINE RL}

We present additional experimental results in the Offline-to-Online (O2O) RL setting, specifically utilizing a pure offline RL algorithm (IQL \citep{kostrikov2021offline}), during the online fine-tuning phase. \Cref{tab:d4rl_iql} summarizes the normalized average returns across various D4RL environments. The results demonstrate that ARB consistently achieves competitive or superior performance compared to other baseline sampling strategies.

\begin{table*}[ht!]
\caption{Normalized average returns on various D4RL environments with IQL.}
\label{tab:d4rl_iql}
\centering
\resizebox{\linewidth}{!}{
\begin{tabular}{l|c|c|c|c|c}
\toprule
\textbf{Environment} & \textbf{Naive} & \textbf{Parallel} & \textbf{Top-N} & \textbf{BERB} & \textbf{ARB(Ours)} \\
\midrule
\multicolumn{6}{l}{\textit{Locomotion Random}} \\
\midrule
halfcheetah-random-v2 & 53.50 $\pm$ 3.19 & 57.90 $\pm$ 2.81 & 60.90 $\pm$ 2.76 & 51.82 $\pm$ 0.72 & 60.40 $\pm$ 2.92 \\
hopper-random-v2 & 15.40 $\pm$ 1.73 & 22.11 $\pm$ 2.58 & 11.40 $\pm$ 1.71 & 10.30 $\pm$ 1.05 & 54.40 $\pm$ 27.35 \\
walker2d-random-v2 & 14.20 $\pm$ 5.43 & 10.80 $\pm$ 3.25 & 13.10 $\pm$ 4.31 & 15.45 $\pm$ 7.55 & 20.80 $\pm$ 7.23 \\
\midrule
\textbf{Average} & 27.70 $\pm$ 3.45 & 30.27 $\pm$ 2.88 & 28.47 $\pm$ 2.93 & 25.86 $\pm$ 3.11 & 45.13 $\pm$ 12.50 \\
\midrule
\multicolumn{6}{l}{\textit{Locomotion Medium Replay}} \\
\midrule
halfcheetah-medium-replay-v2 & 52.90 $\pm$ 1.30 & 49.10 $\pm$ 2.24 & 58.10 $\pm$ 1.93 & 57.40 $\pm$ 1.50 & 62.60 $\pm$ 1.49 \\
hopper-medium-replay-v2 & 97.90 $\pm$ 3.65 & 100.20 $\pm$ 4.09 & 107.81 $\pm$ 4.14 & 103.81 $\pm$ 4.93 & 107.70 $\pm$ 3.25 \\
walker2d-medium-replay-v2 & 98.90 $\pm$ 1.89 & 95.63 $\pm$ 4.29 & 108.40 $\pm$ 2.96 & 98.60 $\pm$ 4.45 & 106.00 $\pm$ 9.20 \\
\midrule
\textbf{Average} & 83.23 $\pm$ 2.28 & 81.64 $\pm$ 3.54 & 91.44 $\pm$ 3.01 & 86.60 $\pm$ 3.63 & 92.10 $\pm$ 4.65 \\
\midrule
\multicolumn{6}{l}{\textit{Locomotion Medium}} \\
\midrule
halfcheetah-medium-v2 & 59.20 $\pm$ 2.87 & 64.40 $\pm$ 4.80 & 70.10 $\pm$ 1.63 & 61.20 $\pm$ 4.90 & 71.30 $\pm$ 2.77 \\
hopper-medium-v2 & 84.40 $\pm$ 5.09 & 103.80 $\pm$ 2.68 & 104.30 $\pm$ 0.78 & 100.10 $\pm$ 4.54 & 109.30 $\pm$ 1.70 \\
walker2d-medium-v2 & 92.70 $\pm$ 3.51 & 94.80 $\pm$ 4.22 & 93.10 $\pm$ 4.29 & 92.20 $\pm$ 6.13 & 108.60 $\pm$ 5.46 \\
\midrule
\textbf{Average} & 78.77 $\pm$ 3.82 & 87.67 $\pm$ 3.90 & 89.20 $\pm$ 2.23 & 84.50 $\pm$ 5.19 & 96.40 $\pm$ 3.31 \\
\midrule
\multicolumn{6}{l}{\textit{Locomotion Medium-Expert}} \\
\midrule
halfcheetah-medium-expert-v2 & 95.03 $\pm$ 2.95 & 95.00 $\pm$ 1.29 & 95.04 $\pm$ 2.06 & 94.30 $\pm$ 3.14 & 95.10 $\pm$ 5.81 \\
hopper-medium-expert-v2 & 109.60 $\pm$ 5.86 & 101.91 $\pm$ 7.82 & 101.62 $\pm$ 6.39 & 111.00 $\pm$ 6.07 & 111.40 $\pm$ 4.39 \\
walker2d-medium-expert-v2 & 118.30 $\pm$ 1.54 & 110.40 $\pm$ 4.70 & 120.10 $\pm$ 6.77 & 116.40 $\pm$ 5.25 & 117.20 $\pm$ 4.92 \\
\midrule
\textbf{Average} & 107.64 $\pm$ 3.45 & 102.44 $\pm$ 4.60 & 105.59 $\pm$ 5.07 & 107.23 $\pm$ 4.82 & 107.90 $\pm$ 5.04 \\
\midrule
\multicolumn{6}{l}{\textit{Antmaze}} \\
\midrule
antmaze-umaze-v2 & 96.00 $\pm$ 1.41 & 96.25 $\pm$ 2.22 & 96.50 $\pm$ 3.87 & 96.50 $\pm$ 1.29 & 97.00 $\pm$ 1.15 \\
antmaze-umaze-diverse-v2 & 80.25 $\pm$ 7.59 & 91.25 $\pm$ 2.87 & 92.50 $\pm$ 1.29 & 96.50 $\pm$ 1.29 & 92.75 $\pm$ 8.02 \\
antmaze-medium-play-v2 & 95.50 $\pm$ 2.65 & 93.00 $\pm$ 5.60 & 91.75 $\pm$ 2.50 & 91.00 $\pm$ 3.46 & 95.00 $\pm$ 3.37 \\
antmaze-medium-diverse-v2 & 88.25 $\pm$ 5.44 & 83.50 $\pm$ 11.03 & 88.50 $\pm$ 1.73 & 94.75 $\pm$ 1.50 & 91.75 $\pm$ 2.63 \\
antmaze-large-play-v2 & 73.00 $\pm$ 6.98 & 70.00 $\pm$ 6.68 & 70.75 $\pm$ 3.86 & 70.25 $\pm$ 8.38 & 76.00 $\pm$ 4.69 \\
antmaze-large-diverse-v2 & 66.25 $\pm$ 8.66 & 71.00 $\pm$ 5.03 & 69.75 $\pm$ 4.19 & 72.50 $\pm$ 7.94 & 76.00 $\pm$ 8.04 \\
\midrule
\textbf{Average} & 83.21 $\pm$ 11.52 & 84.17 $\pm$ 11.23 & 84.96 $\pm$ 11.45 & 86.92 $\pm$ 12.06 & 88.08 $\pm$ 8.87 \\
\bottomrule
\end{tabular}
}
\end{table*}

\end{document}